%% file: main.tex
\documentclass[10pt,twocolumn,letterpaper]{article}

\usepackage[pagenumbers]{cvpr} 

\input{preamble}

\definecolor{cvprblue}{rgb}{0.21,0.49,0.74}
\usepackage[pagebackref,breaklinks,colorlinks,citecolor=cvprblue]{hyperref}
\usepackage[super]{nth}
\usepackage{nicefrac}
\usepackage{array}
\usepackage{multirow}

\def\paperID{16054} 
\def\confName{CVPR}
\def\confYear{2024}

\title{Neural Markov Random Field for Stereo Matching}

\author{Tongfan Guan$^1$\\
{\tt\small tfguan@link.cuhk.edu.hk}
\and
Chen Wang$^2$\\
{\tt\small chenw@sairlab.org}
\and
Yun-Hui Liu$^1$\thanks{Corresponding author: Yun-Hui Liu}\\
{\tt\small yhliu@mae.cuhk.edu.hk}
\and
$^1$The Chinese University of Hong Kong
\and
$^2$\href{https://sairlab.org}{\color{black}Spatial AI \& Robotics Lab}, University at Buffalo
}

\begin{document}
\maketitle
\input{0_abstract}    
\input{1_intro}
\input{2_relatedworks}
\input{3_methodology}
\input{4_experiment}
{
    \small
    \bibliographystyle{ieeenat_fullname}
    \bibliography{main}
}

\input{X_suppl}

\end{document}

%% file: preamble.tex
\usepackage[table,dvipsnames]{xcolor}


%% file: 0_abstract.tex
\begin{abstract}
Stereo matching is a core task for many computer vision and robotics applications. 
Despite their dominance in traditional stereo methods, the hand-crafted Markov Random Field (MRF) models lack sufficient modeling accuracy compared to end-to-end deep models. 
While deep learning representations have greatly improved the unary terms of the MRF models, the overall accuracy is still severely limited by the hand-crafted pairwise terms and message passing. 
To address these issues, we propose a neural MRF model, where both potential functions and message passing are designed using data-driven neural networks. 
Our fully data-driven model is built on the foundation of variational inference theory, to prevent convergence issues and retain stereo MRF's graph inductive bias. 
To make the inference tractable and scale well to high-resolution images, we also propose a Disparity Proposal Network (DPN) to adaptively prune the search space of disparity. 
The proposed approach ranks \nth{1} on both KITTI 2012 and 2015 leaderboards among all published methods while running faster than 100 ms. 
This approach significantly outperforms prior global methods, \eg, lowering D1 metric by more than 50\% on KITTI 2015. 
In addition, our method exhibits strong cross-domain generalization and can recover sharp edges. The codes at \url{https://github.com/aeolusguan/NMRF}.
\end{abstract}

%% file: 1_intro.tex
\section{Introduction}
\label{sec:intro}

Stereo matching is a critical component in computer vision, mimicking human binocular vision to cognize 3D information within the field of view~\cite{scharstein2002taxonomy}. 
Given a pair of images, it aims to determine the horizontal displacement of individual pixels in one image to align with the other. 
Stereo matching has been applied to many fields like 3D reconstruction~\cite{tanner2022large}, autonomous navigation~\cite{wang2019pseudo}, and augmented reality~\cite{zenati2007dense}, bridging the gap between digital imagery and real-worlds.

Among the various methodologies employed in stereo matching, Markov Random Field (MRF)~\cite{tappen2003comparison} stands out as one of the most widely used and effective models.
MRFs leverage a probabilistic model to explain observed image features and enforce spatial coherence, producing piecewise smooth disparity maps.
Due to their ability to reduce matching ambiguities in challenging regions~\cite{szeliski2008comparative}, MRF and its variants~\cite{birchfield1999multiway,hirschmuller2007stereo,zhang2015meshstereo,Taniai18} have dominated the field before deep neural networks~\cite{zbontar2016stereo,kendall2017end, lipson2021raft} emerged, according to Middlebury \cite{scharstein2002taxonomy}, a popular benchmark for stereo matching.

\begin{figure}[!t]
  \centering
  \includegraphics[width=0.98\linewidth]{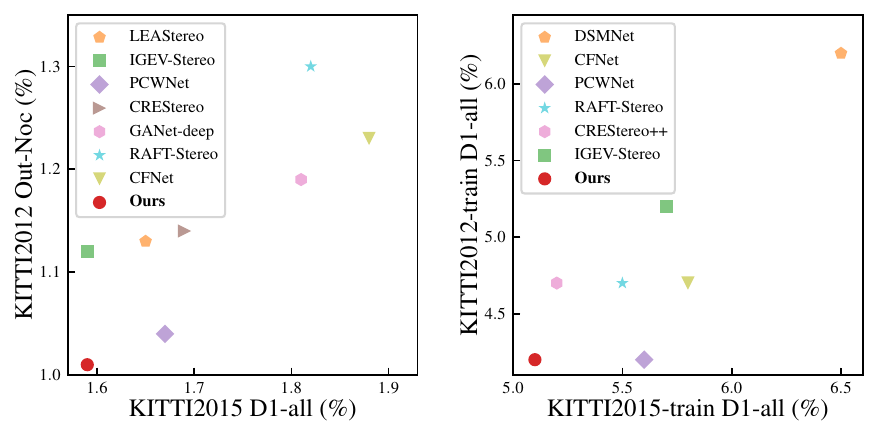}\\
  \vspace{-0.4em}
  \makebox[0.58\linewidth]{\footnotesize (a)}
  \makebox[0.4\linewidth]{\footnotesize (b)}
  \vspace{-0.6em}
  \caption{(a) Comparison with state-of-the-art stereo methods~\cite{cheng2020hierarchical,xu2023iterative,shen2022pcw,li2022practical,Zhang2019GANet,lipson2021raft,Shen_2021_CVPR} on KITTI 2012 and 2015 leaderboards. (b) Cross-domain generalization comparison with current robust methods~\cite{zhang2019domaininvariant,Shen_2021_CVPR,shen2022pcw,lipson2021raft,jing2023uncertainty,xu2023iterative}. All methods are only trained on the synthetic SceneFlow dataset~\cite{mayer2016large} and evaluated on KITTI2012/2015 trainsets with fixed parameters.}
  \label{fig:scatter}
  \vspace{-1.2em}
\end{figure}

Although MRFs have achieved promising results, they are still often faced with difficulties because of hand-crafted \textit{potential functions} and \textit{message passing} procedures.
Typically, an MRF's potential function comprises a \textit{unary} term that evaluates the similarity in intensity/gradient of matching pixels/features; and a \textit{pairwise} term that penalizes solutions that violate certain spatial smoothness criteria~\cite{szeliski2008comparative}.  
However, such hand-crafted approaches cannot fully model all scenarios, e.g., carefully designed models for object occlusion may not be able to model abrupt changes in disparity at object boundaries. This inadequacy often results in an over-smoothed disparity map.
Moreover, hand-crafted message passing may also struggle to handle complex pairwise relationships. This limitation can often lead to difficulties such as inaccurate disparity estimations or convergence issues. 
Despite the unary terms have been greatly improved by deep feature representations~\cite{chen2015deep,luo2016efficient,knobelreiter2017end,slossberg2016deep,knobelreiter2020belief}, the overall accuracy of MRFs is still severely limited by the hand-crafted pairwise potential and message passing functions.

To tackle the above problems, we propose a Neural MRF (NMRF) model to learn the complicated pixel relationships in a data-driven manner, mitigating the inefficacy of manual design.
The mean-field variational inference theory is leveraged to design neural modules that perform as unary/pairwise potential terms and message passing.
Additionally, to make our NMRF tractable and scale well to high-resolution images, we propose a Disparity Proposal Network (DPN) which significantly prunes the search space of disparity with little sacrifice of performance.
To the best of our knowledge, NMRF is the first fully data-driven stereo MRF model while retaining its strong graph inductive bias to handle uncertainty and ambiguity in image data.

\begin{figure*}[!t]
  \centering
  \begin{subfigure}[t]{0.63\linewidth}
    \includegraphics[width=\linewidth]{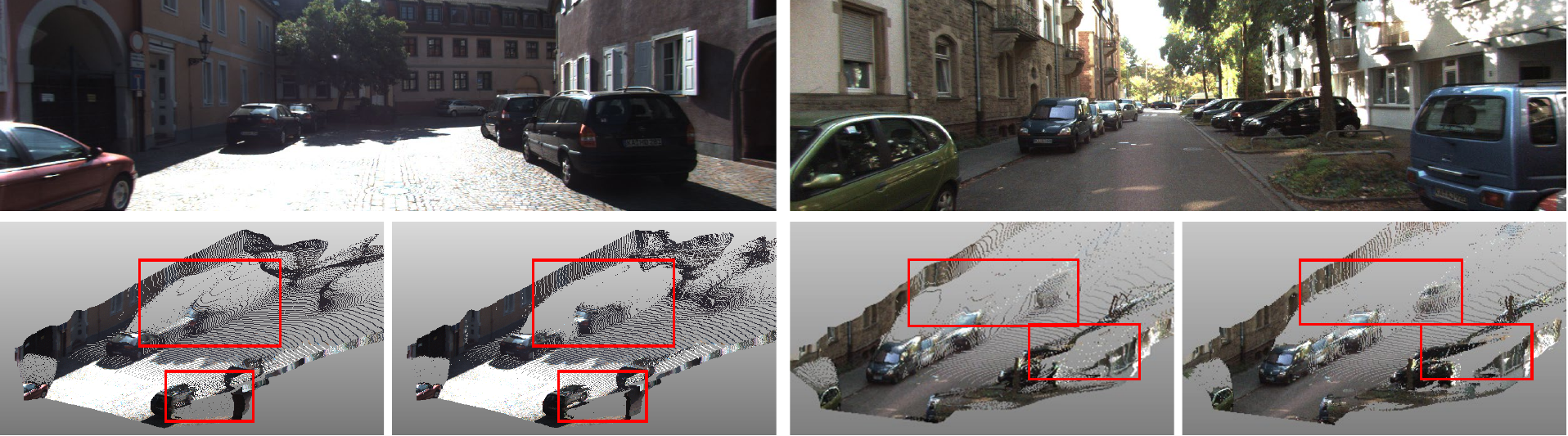}\\
    \makebox[0.245\linewidth]{\footnotesize LEAStereo~\cite{cheng2020hierarchical}}
    \makebox[0.245\linewidth]{\footnotesize Ours}
    \makebox[0.245\linewidth]{\footnotesize LEAStereo~\cite{cheng2020hierarchical}}
    \makebox[0.245\linewidth]{\footnotesize Ours}
    \caption{}
    \label{fig:kitti-pt}
  \end{subfigure}
  \hfill
  \begin{subfigure}[t]{0.352\linewidth}
    \includegraphics[width=\linewidth]{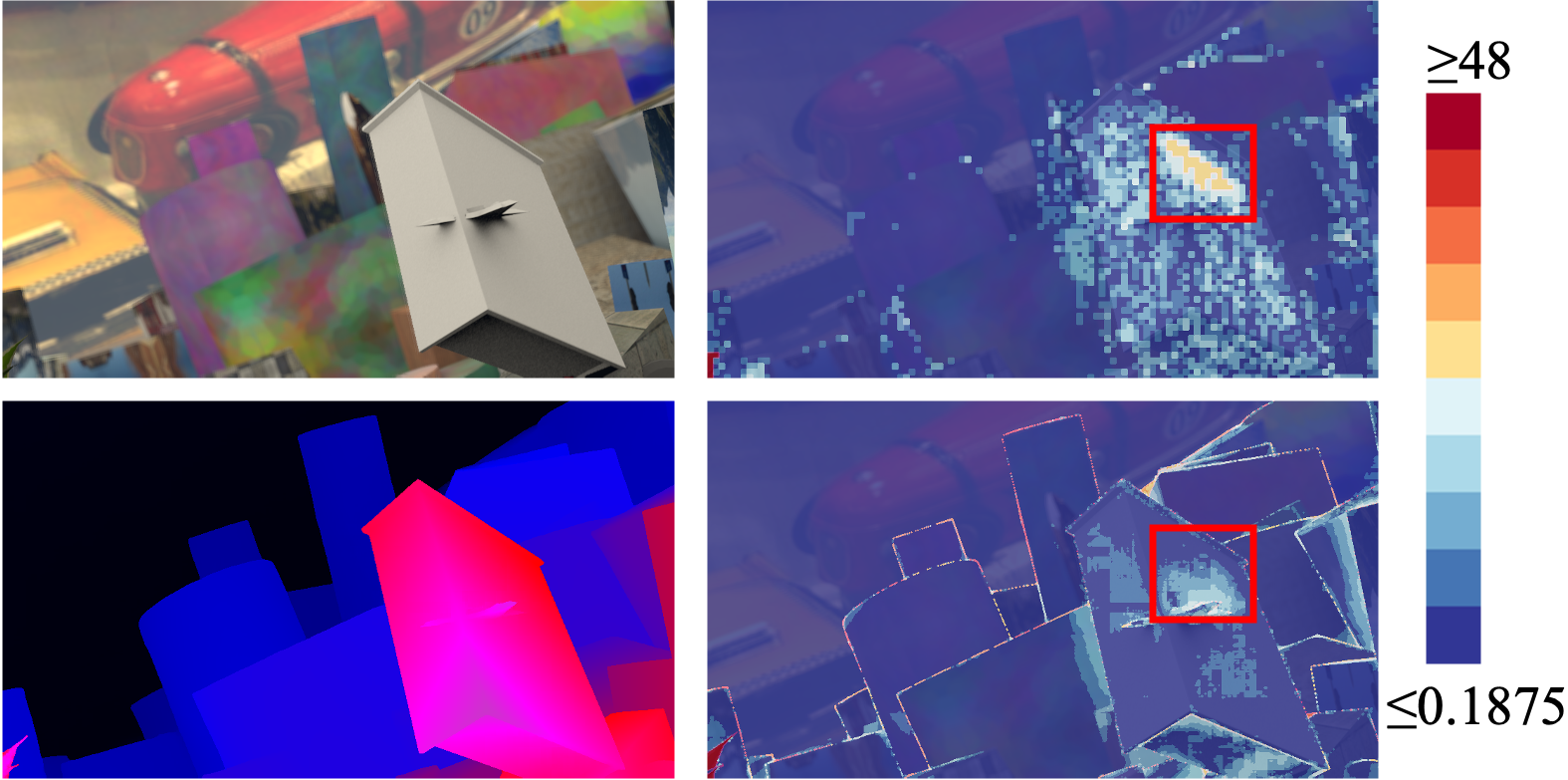}\\
    \makebox[0.43\linewidth]{\footnotesize Left \& Disparity}
    \makebox[0.43\linewidth]{\footnotesize Error map}
    \caption{}
    \label{fig:scene-robust}
  \end{subfigure}
  \vspace{-0.6em}
  \caption{(a) Stereo point cloud comparison between LEAStereo~\cite{cheng2020hierarchical} and our method on KITTI test set. Notice how our approach notably alleviates flying pixels near object boundaries, which is well-known as over-smoothing problem~\cite{chen2019over}. Please zoom in for more details. (b)  Left column: left image (top) and disparity estimation (bottom), Right column: color-coded error map of pixelwise best proposal (top) and disparity estimation (bottom). This method even recovers from proposal failure (marked with red) in the large textureless region.}
  \vspace{-1.2em}
\end{figure*}

Our NMRF model reports state-of-the-art accuracy on SceneFlow~\cite{mayer2016large} and ranks \nth{1} on KITTI 2012~\cite{geiger2012we} and 2015~\cite{menze2015object} leaderboards among all published methods while running faster than 100 ms. Compared with previous global stereo networks, NMRF outperforms with a substantial margin, \eg, reducing the \textit{D1-bg} outlier ratio by more than 50\% on KITTI 2015, 1.28\% (Ours) \vs 2.85\% (LBPS~\cite{knobelreiter2020belief}). NMRF also exhibits state-of-the-art cross-domain generalization ability. When trained only on synthetic SceneFlow dataset, NMRF performs very well on real datasets KITTI~\cite{menze2015object,geiger2012we}, Middlebury~\cite{scharstein2014high}, and ETH3D~\cite{schops2017multi}. Furthermore, NMRF is able to recover sharp depth boundaries, as shown in \cref{fig:kitti-pt}, which is key to downstream tasks, such as 3D reconstruction and object detection. 

In summary, our contributions include:

\begin{itemize}
    \item We introduce a novel fully data-driven MRF model for stereo matching that can effectively learn complicated relationships between pixels from data.
    \item We develop a search space pruning module that largely reduces the computation load of neural MRF inference, which is also valuable in other dense matching tasks.
    \item Our architecture achieves state-of-the-art results on popular benchmarks in terms of both accuracy and robustness.
\end{itemize}


%% file: 2_relatedworks.tex
\section{Related Work}
\label{sec:related}

Since MC-CNN~\cite{zbontar2016stereo}, deep learning has been leveraged for unary matching cost computation~\cite{zbontar2016stereo,luo2016efficient,knobelreiter2017end,chen2015deep,han2015matchnet,zagoruyko2015learning}, cost volume aggregation~\cite{kendall2017end,chang2018pyramid,guo2019group,Zhang2019GANet,zhang2019domaininvariant,cheng2020hierarchical,Shen_2021_CVPR,shen2022pcw,xu2022attention,cheng2019learning,xu2020aanet}, and iterative disparity refinement~\cite{kendall2017end,mayer2016large,Liang2018Learning,khamis2018stereonet,tulyakov2018practical,xu2020aanet,chabra2019stereodrnet,tankovich2021hitnet,lipson2021raft,li2022practical,jing2023uncertainty,xu2023iterative}. 
Laga \etal~\cite{laga2020survey} give a thoroughly survey on deep techniques for stereo matching. This section focuses on MRF-related stereo networks.

\paragraph{Stereo MRF networks.} MRFs formulate stereo matching as a pixel-labeling problem, and assign every pixel $o$ a disparity label $z_o$. 
The set of all pixel-label assignments is denoted by $\{z_o\}$. We write MRFs' probabilistic model as:
\begin{equation}
    p(\{z_o\},\{\mathbf{x}_o\})\propto \prod_o \Phi(z_o,\mathbf{x}_o) \prod_{(o,p)} \Psi(z_o,z_p),
    \label{eq:mrf-tra}
\end{equation}
where $\Phi$ and $\Psi$ are non-negative unary and pairwise \textit{potential functions}, respectively; $(o,p)$ represents a pair of neighboring pixels; and $\mathbf{x}_o$ is the observed pixel features. 
The Maximum A Posterior (MAP) estimate of $\{z_o\}$ is equivalent to energy minimization by taking \textit{log} of \cref{eq:mrf-tra}. 

Typically, $\Phi$ and $\Psi$ are hand-crafted based on stereo domain knowledge, \eg, intensity constancy and piecewise coplanar~\cite{Taniai18,bleyer2011patchmatch}. 
The pioneering MC-CNN~\cite{zbontar2016stereo} generalizes manually designed unary potential $\Phi$ with a siamese network, which performs feature extraction from image patches and computes unary costs based on a fully-connected DNN. 
Chen \etal~\cite{chen2015deep} achieved 100$\times$ speed-up compared to MC-CNN by replacing the fully-connected DNN with a dot product layer at the cost of little performance drop. 
Instead of independent predictions on pairs of image patches, Luo \etal~\cite{luo2016efficient} compared a patch in the left image with a horizontal stripe in the right image to extract marginal distributions over all possible disparities. 
However, these methods still leverage the hand-crafted pairwise potential and message passing of  SGM~\cite{hirschmuller2007stereo}. 
Considering the difficulty of tuning SGM parameters $(P_1$, $P_2)$ to accurately penalize disparity discontinuities for different cases, Seki \etal~\cite{seki2017sgm} trained a neural network to provide adaptive parameters $P_1(o)$ and $P_2(o)$ for every pixel $o$.
Similarly, PBCP~\cite{seki2016patch} set them based on the disparity confidence map estimated with a CNN. 
Hybrid CNN-CRF~\cite{knobelreiter2017end} performed feature matching on complete images, and this setting firstly enabled end-to-end joint learning of SGM message passing and the unary/pairwise CNNs. 
Moreover, GANet~\cite{Zhang2019GANet} proposed a differentiable approximation of SGM by replacing user-defined parameters with learned guidance weight matrix.
Recently, LBPS~\cite{knobelreiter2020belief} adapted Belief Propagation (BP) to learning formulation through a differentiable loss defined on marginal distributions, making graphical models fully compatible with deep learning. 
However, the application of hand-crafted message passing functions didn't achieve top performance\footnote{The overall accuracy of GANet~\cite{Zhang2019GANet} mainly owing to 3D convolutional cost aggregation. Approximate SGM aggregation alone didn't achieve satisfactory accuracy, as evidenced by its ablation studies.} due to its limited ability to handle complicated pairwise relationships.

\paragraph{Neural message passing.} Gilmer \etal~\cite{gilmer2017neural} demonstrated that neural message passing (NMP) is the cornerstone of recent successful models on graph-structured data. 
Specifically, it exchanges vector messages between nodes and updates node embeddings using neural networks. 
In computer vision tasks, it has been employed to exchange descriptor information among keypoints in sparse feature matching~\cite{sarlin2020superglue,sun2021loftr} and solve the maximum network flow formulation of multiple object tracking~\cite{braso2020learning}. 
Dai \etal~\cite{dai2016discriminative} noted that neural message passing can be derived through certain embedded inference on probabilistic graphical models (PGM)~\cite{hamilton2020graph}. 
This observation motivates us to build our neural message passing function on the foundation of variational inference theory, to prevent convergence issues of MRF optimization.
Given that Transformer~\cite{liu2021Swin,dong2022cswin} is also a form of message passing technique, we incorporate key elements of Transformer into our message passing function while retaining the graph inductive bias of stereo MRF.

\paragraph{Label space pruning.} Many algorithms devoted to pruning MRF label space due to tractability or efficiency reason. 
Menze \etal~\cite{menze2015discrete} pruned two-dimensional flow space using descriptor matching. 
To prune the disparity search space, DeepPruner~\cite{duggal2019deeppruner} designed a differentiable PatchMatch-based pruner to predict a confidence range for each pixel. 
CFNet~\cite{Shen_2021_CVPR} and UCSNet~\cite{cheng2020deep} gradually narrowed down the disparity search space in the cascaded multi-scale manner, where next scale's search range is generated based on the disparity estimate of current scale. 
The unimodal search range $[l_o,r_o]$ outputted by these popular methods~\cite{duggal2019deeppruner,Shen_2021_CVPR,cheng2020deep} for each pixel $o$, however, may be susceptible to local optimum. 
In contrast, the proposed Disparity Proposal Network (DPN) has the ability to predict multi-modal proposals.

%% file: 3_methodology.tex
\section{Methodology}\label{sec:impl}
\begin{figure*}[!t]
  \centering
  \includegraphics[width=0.96\linewidth]{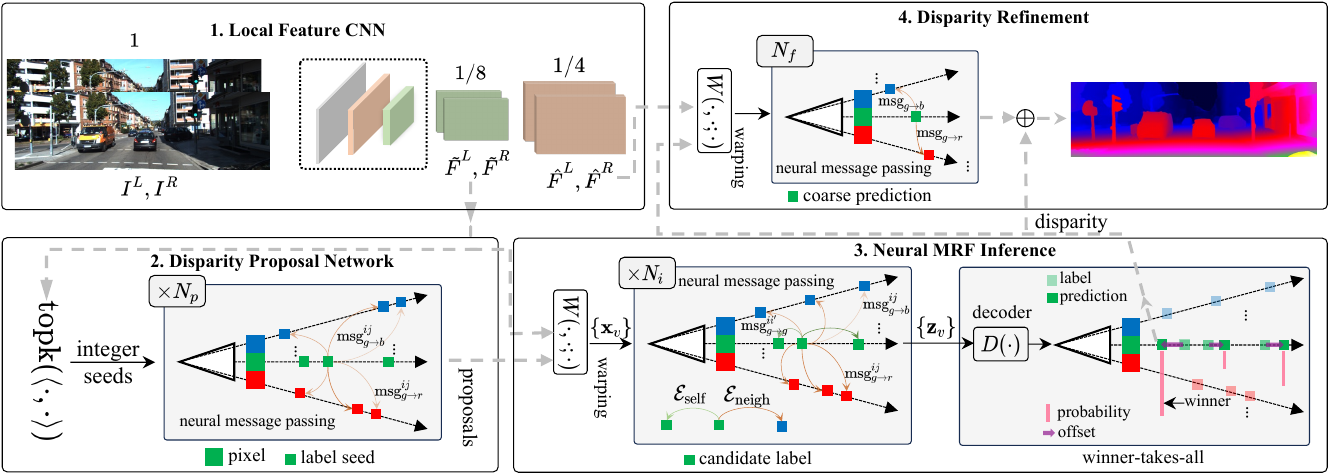}
  \vspace{-0.3em}
  \caption{\textbf{Overview of the proposed method.} It has four components: \textbf{1.} A local feature CNN extracts the coarse and fine-level feature maps from the input image pair.
  \textbf{2.} A disparity proposal network prunes space of disparity. For every pixel, the top $k$ disparity modals are identified, and then updated using $N_p$ neural message passing, resulting in a sparse label set $L_o$.
  \textbf{3.} The MRF factorizes into a probabilistic graph, where each node corresponds to a candidate label and each edge connects a label pair from neighbor pixels. Different potential functions are used for intra- and inter-pixel label pairs respectively. The inferred latent embedding $\mathbf{z}_v$ is then decoded to posterior probability and offset. The winner label is selected as the coarse prediction.
  \textbf{4.} Disparity refinement also leverages a neural MRF model but with only one label per pixel for efficiency. The inferred latent embeddings are decoded into disparity residuals.}
  \label{fig:overview}
  \vspace{-1.2em}
\end{figure*}

Given an image pair $(I^L, I^R)$, we propose to estimate the disparity map via a Neural Markov Random Field (NMRF) model. 
An overview of the approach is presented in \cref{fig:overview}.

This approach consists of two stages. 
The first stage employs a NMRF model to infer disparities on the coarse level ($\nicefrac{1}{8}$) features (\cref{sec:inference}). 
To make NMRF efficient, a Disparity Proposal Network is proposed to prune candidate space (\cref{sec:dpn}). 
The next stage performs disparity refinement on the fine level ($\nicefrac{1}{4}$) features (\cref{sec:refinement}). 
A local feature CNN provides the coarse and fine level features for both stages.

\paragraph{Local feature CNN.} To extract multi-level features from the pair of images, we employ a siamese convolutional network, which comprises a stack of residual blocks, instance normalization and downsampling layers. 
The coarse level features at $\nicefrac{1}{8}$ of the original image resolution are denoted as $\tilde{F}^L$ and $\tilde{F}^R$, while the fine-level features at $\nicefrac{1}{4}$ of the original image resolution are denoted as $\hat{F}^L$ and $\hat{F}^R$.

\subsection{Neural MRF Formulation}
\label{sec:mrf-model}
We start by formulating the NMRF model for stereo matching before implementing it in \cref{sec:inference}. 
Given a discrete candidate label (disparity) set $L_o\subset\mathbb{R}^+$ for every pixel $o$, hand-crafted MRFs center on a carefully designed energy function over pixel-label assignments $\{z_o|z_o\in L_o\}$. 
However, the pairwise terms typically depend only on label difference $|z_o-z_p|$ while overlooking image content~\cite{szeliski2008comparative}. 
In order to incorporate additional information, such as 3D geometry and visual context, into potential functions, our NMRF represents each label with an \textit{embedding} $\mathbf{z}_v\in\mathbb{R}^d$. 

A graph $\mathcal{G}=(\mathcal{V},\mathcal{E})$ defines the discrete neural MRF as:
\begin{equation}
    p(\{\mathbf{x}_v\},\{\mathbf{z}_v\}) \propto \prod_{v\in\mathcal{V}} \Phi(\mathbf{x}_v,\mathbf{z}_v)\prod_{(u,v)\in\mathcal{E}}\Psi(\mathbf{z}_u,\mathbf{z}_v),\\
\end{equation}
where each node $v$ corresponds to a candidate label, and the edge $(u,v)$ connects a pair of labels from neighbor pixels within a $M\times M$ image window. 
Intuitively, $\Phi(\mathbf{x}_v,\mathbf{z}_v)$ indicates the data likelihood of observed label feature $\mathbf{x}_v$ given its latent embedding $\mathbf{z}_v$, while $\Psi(\cdot,\cdot)$ controls the penalty of pairwise label assignments. 
Given observed label features $\{\mathbf{x}_v\}$, our goal is to infer the latent embeddings $\{\mathbf{z}_v\}$, from which the disparity estimation will be deduced.

As shown in the \nth{3} part of \cref{fig:overview}, this graph has two types of edges. Intra-pixel edges, or \textit{self} edges, $\mathcal{E}_{\text{self}}$, connect labels $v$ to all other labels of the same pixel. 
Inter-pixel edges, or \textit{neighbor} edges, $\mathcal{E}_{\text{neigh}}$, connect labels $v$ to all labels of neighbor pixels. 
Self edges are usually overlooked in hand-crafted stereo MRFs. 
In contrast, this paper accounts for self edges either, but uses different potential functions, \ie, $\Psi_{\text{neigh}}$ and $\Psi_{\text{self}}$, for the two types of edges. 
Intuitively, $\Psi_{\text{neigh}}$ allows compatible label pairs to support each other, whereas $\Psi_{\text{self}}$ expects labels of the same pixel to compete with each other. 
\Cref{tab:ablation_comp} shows that this inductive bias for pairwise potential will bring considerable improvements.

In hand-crafted MRF settings, $\Phi$ and $\Psi$ are typically designed based on stereo domain knowledge. 
In our neural MRF, however, we do not specify the exact form of potential functions $\Phi$ and $\Psi$. 
Instead, we will seek to implicitly learn these potential functions by inferring latent embeddings $\{\mathbf{z}_v\}$ that can explain the ground truth disparity map.

\paragraph{Observed label feature.} The observed feature of a candidate label must integrate matching cues from both left and right views. 
Given the coarse level features $\tilde{F}^L$ and $\tilde{F}^R$, we compute the observed feature $\mathbf{x}_v$ of a candidate label positioned at $\mathbf{p}_v:=(i,j,z)$ using a warping function \wrt $\tilde{F}^L(i,j)$ and $\tilde{F}^R(i-z,j)$. 
We leave the formal definition of the warping function in the supplementary material.

\subsection{Neural MRF Inference}
\label{sec:inference}
\noindent\textbf{Preliminaries: Embedding mean-field inference~\cite{dai2016discriminative,hamilton2020graph}.}
Exact computation of posterior $p(\{\mathbf{z}_v\}|\{\mathbf{x}_v\})$ is intractable, even if $\Phi$ and $\Psi$ are known and well-defined. 
The mean-field theory assumes that the posterior over latent variables can be factorized into $\mathcal{V}$ independent marginals $q_v({\mathbf{z}_v}$), \ie, $p(\{\mathbf{z}_v\}|\{\mathbf{x}_v\}) \approx \prod_v q_v(\mathbf{z}_v)$.
The optimal marginals $\{q_v\}$ under the mean-field assumption are obtained by minimizing the Kullback-Leibler (KL) divergence between the approximation posterior and the true posterior. 
Hamilton~\cite{hamilton2020graph} shows that the optimal solution needs to satisfy the following fixed point equation for all $v\in\mathcal{V}$:
\begin{multline}
  \log(q_v(\mathbf{z}_v))=c_v + \log(\Phi(\mathbf{x}_v,\mathbf{z}_v)) + \\\sum_{u\in\mathcal{N}(v)}\int_{\mathbb{R}^d}q_u(\mathbf{z}_u)\log(\Psi(\mathbf{z}_u,\mathbf{z}_v))\mathrm{d}\mathbf{z}_u,
  \label{eq:fixed-point}
\end{multline}
where $c_v$ is a normalization constant and $\mathcal{N}(v)$ denotes the neighbor set of node $v$. 
It implies that marginal distribution $q_v(\mathbf{z}_v)$ is a function of node feature $\mathbf{x}_v$ and neighbor marginals $q_u(\mathbf{z}_u),\forall u\in\mathcal{N}(v)$, \ie, $q_v(\mathbf{z}_v)=f(\mathbf{z}_v,\mathbf{x}_v,\{q_u\}_{u\in\mathcal{N}(v)})$.  
Supposing an injective feature map $\phi$, each marginal $q_v(\mathbf{z}_v)$ maps to an embedding $\mu_v=\int_{\mathbb{R}^d}q_v(\mathbf{z}_v)\phi(\mathbf{z}_v)\mathrm{d}\mathbf{z}_v \in \mathbb{R}^d$. 
The solution of fixed point \cref{eq:fixed-point} in the embedding space could be approximated by iteratively evaluating
\begin{equation}
    \mu_v^t=\tilde{f}(\mathbf{x}_v,\mu_v^{t-1},\{\mu_u^{t-1}\}_{u\in\mathcal{N}(v)}),
    \label{eq:em-update}
\end{equation}
where $\tilde{f}$ is a vector-valued function and has complicated nonlinear dependencies on the potential functions $\Phi$, $\Psi$, as well as the feature map $\phi$. 

\paragraph{Inference with neural message passing.} Currently, mean-field inference is only tractable for restrictive potential functions, \eg, those from the exponential family. 
In our setting, it's hard to work out $\tilde{f}$ in \cref{eq:em-update} since potential functions $\Phi$, $\Psi$, and the feature map $\phi$ are unknown and need to be learned from data. 
To address this dilemma, we  parameterize $\tilde{f}$ to be a neural network that retains the inductive bias of mean-field inference. 
Notice that $\tilde{f}$ exactly corresponds to some message passing operation as it \textit{aggregates} information from neighbor embeddings (\ie, $\{\mu_u\}_{u\in\mathcal{N}(v)}$) and \textit{updates} the node's current representation (\ie, $\mu_v^{t-1}$).

The initial representation $^{(0)}\mu_v$ for label $v$ is initialized with $\mathbf{x}_v$. 
Let $^{(\ell)}\mu_v$ be the intermediate embedding for label $v$ at layer $\ell$. 
The message $\mathbf{m}_{\mathcal{E}^\ell\rightarrow v}$ is the outcome of embedding aggregation along edges $\mathcal{E}^\ell $, where $\mathcal{E}^\ell\in\{\mathcal{E}_{\text{self}},\mathcal{E}_{\text{neigh}}\}$. The message passing update is formally defined as:
\begin{equation}
\begin{split}
  ^{(\ell+1)} \hat{\mu}_v& = {^{(\ell)}\mu_v} + \mathbf{m}_{\mathcal{E}^\ell\rightarrow v}\\
  ^{(\ell+1)}\mu_v&=\text{MLP}\left(^{(\ell+1)}\hat{\mu}_v\right) + {^{(\ell+1)}\hat{\mu}_v},
\end{split}
\end{equation}
where the MLP conceptually corresponds to the unary potential function $\Phi$ since it controls the embedding update, as shown in \cref{eq:fixed-point}. 
Furthermore, we observe that potential function $\Psi$ controls the weight to aggregate information from neighbor embeddings. 
Intuitively, pairwise potential $\Psi(\cdot,\cdot)$ measures \textit{affinity} between connected labels. 
This observation, together with the success of Transformer, motivates us to leverage self-attention for the embedding aggregation. 
A fixed number of $N_i$ layers are chained and alternatively aggregate along the neighbor and self edges.

\paragraph{Attentional aggregation.} To aggregate embedding along the neighbor edges $\mathcal{E}_{\text{neigh}}$, a label $v$ first extracts relevant information from all neighbors $\{u:(u,v)\in\mathcal{E}_{\text{neigh}}\}$, and then sums them up weighted by the affinity score. 
The message computation could be formulated as:
\begin{equation}
\begin{split}
 &\mathbf{m}_{\mathcal{E}_{\text{neigh}\rightarrow v}}=\sum_{(u,v)\in\mathcal{E}_{\text{neigh}}} \alpha_{uv} (\mathbf{v}_u+\mathbf{r}_{u-v}^\mathrm{v})
  \\
 &\alpha_{uv}=\text{softmax}_u (\mathbf{q}_v^\top\mathbf{k}_u+\underbrace{\mathbf{q}_v^\top\mathbf{r}_{u-v}^\mathrm{k} + \mathbf{k}_u^\top\mathbf{r}_{u-v}^\mathrm{q})}_{\text{content-adaptive positional bias}},
\end{split}
\label{eq:agg}
\end{equation}
where $\mathbf{q}_v,\mathbf{k}_v,\mathbf{v}_v$ are the \textit{query, key} and \textit{value} get by a linear projection of the embedding $^{(\ell)}\mu_v$, and $\mathbf{r}_{u-v}^\mathrm{q}, \mathbf{r}_{u-v}^\mathrm{k},\mathbf{r}_{u-v}^\mathrm{v}$ are encodings of the relative position $\mathbf{p}_u-\mathbf{p}_v$ in three different subspaces. 
Adding relative positional encoding $\mathbf{r}_{u-v}^\mathrm{v}$ to the value vector, messages will become position-dependent. This will benefit disparity aggregation in ambiguous areas, where the relative position is important. 
The query $\mathbf{q}_v$ and key $\mathbf{k}_u$ could contain information about where to focus in the neighborhood. 
Thus, their dot product with the relative positional encodings are utilized as content-adaptive postional bias of self-attention. 
STTR~\cite{li2021revisiting} has found similar positional bias is beneficial for stereo matching.

The 3D relative position $\mathbf{p}_u-\mathbf{p}_v$ consists of pixel coordinate difference $(\Delta i, \Delta j)$, and disparity difference $\Delta z$. 
In our setting, however, it is memory unaffordable to directly encode $\mathbf{p}_u-\mathbf{p}_v$ in the sinusoidal format or through a small network~\cite{zhao2021point} due to the quadratic pairwise combination. 
Thus, we pursue an approximate encoding method. 
Given that label $v$ and $u$ come from a pair of pixels within a $M\times M$ window, the pixel coordinate difference $(\Delta i, \Delta j)$ will take integer values in the range $[-M+1,M-1]$ along each axis.
With this in mind, we retrieve the position encoding of $(\Delta_i,\Delta_j)$ from a learnable embedding table $\mathbf{P}\in\mathbb{R}^{3\times (2M-1)\times (2M-1)\times D}$, where the leading dimension $3$ is for query, key and value respectively, and $D$ is the number of channels. 
For the encoding of relative disparity $\Delta z$, we approximate with sinusoidal encoding of absolute disparity. 
Specifically, we concatenate it with embedding $^{(\ell)}\mu_v$ before the linear projection to get query, key and value.

For embedding aggregation along self edges $\mathcal{E}_{\text{self}}$, the pixel coordinate difference $(\Delta i, \Delta j)$ are always zero, and thus we omit the related position encoding terms in \cref{eq:agg}.

\paragraph{Disparity estimation.}
After $N_i$ layers' neural message passing, we estimate disparity map by decoding the inferred latent embedding $\{^{(N_i)}\mu_v\}$. 
For every pixel on the coarse level, the latent embedding of each candidate label is decoded into $8\times 8$ disparity offsets and $8\times 8$ probabilities. 
The $8\times 8$ disparity offsets \wrt the candidate label are used to compute disparity hypotheses for the $8\times 8$ pixels in the original image. 
As a result, we produce $k$, the number of candidate labels, scored (with probability) disparity hypotheses for every pixel in the input image. 
We estimate the disparity using the winner-takes-all strategy, as depicted in \cref{fig:overview}.

\subsection{Disparity Proposal Network}
\label{sec:dpn}
To ensure tractable neural MRF inference, we propose a Disparity Proposal Network (DPN) to provide each pixel $o$ with a \textit{small} candidate label space $L_o$. 
The DPN first identifies the top $k$ disparity modals in the range $[0, z_{\text{max}}]$. Then, it updates them by leveraging the inherent spatial coherence, resulting in $k$ candidate labels for each pixel.

\paragraph{Top $k$ label seeds.} At pixel $o=(i,j)$, the initial matching score for integer disparity $z\in [0,z_{\text{max}}]$ is computed using the inner product $\langle\tilde{F}^L(i,j),\tilde{F}^R(i,j-z)\rangle$. 
We identify the disparity modals, \ie, the integer disparities with locally maximum matching scores, using a 1D \textit{max pooling} along the disparity dimension. 
The kernel size is set to $3$ so as to detect the local maximum within the vicinity of $[-1,1]$. 
Then a \verb|topk| operation identifies the best $k$ modals. 
Each modal is characterized by its \textit{position} $\mathbf{p}$ and \textit{matching feature} $\mathbf{d}$ (details in supplementary). 
We refer to them jointly $(\mathbf{p},\mathbf{d})$ as the \textit{label seed}. 
The position consists of $i$ and $j$ coordinates as well as the disparity $z$, \ie, $\mathbf{p}_v:=(i,j,z)$.

\paragraph{Label seeds propagation.} The top $k$ label seeds may fail to capture the true disparities, particularly in the textureless or occluded regions. 
Even so, good label seeds still make up the majority in the entire image field. 
Our intuition is to rectify erroneous label seeds using the information from dependent good label seeds. 
To facilitate this, we also utilize a message passing network to exchange matching features between label seeds. 
It is well-known that long-range dependency is critical for tackling occlusion and large textureless regions. 
To efficiently exchange matching features with distant label seeds, we implement message aggregation using a cross-shaped window self-attention~\cite{dong2022cswin}.
In contrast to neural MRF inference, DPN ignores \textit{self} edges since intra-pixel competition does not make sense in the proposal extraction stage. 
The enhanced matching features of each label seed  is decoded into residuals \wrt the initial integer disparity. More details about the cross-shaped attentional message aggregation are provided in the supplementary material.

\begin{table*}[!t]
\small
\centering
\begin{tabular}{c c c c c c c c}
\toprule 
\textbf{Methods} & PSMNet~\cite{chang2018pyramid} & GANet-deep~\cite{Zhang2019GANet} &  AANet~\cite{xu2020aanet} & 
LEAStereo~\cite{cheng2020hierarchical} & ACVNet~\cite{xu2022attention} & IGEV-Stereo~\cite{xu2023iterative} & Ours\\
\midrule
\textbf{EPE} [px] & 1.09 & 0.78 & 0.87 & 0.78 & 0.48 & 0.47 & \textbf{0.45}\\
\textbf{Bad 1.0} [\%] &12.1 &8.7&9.3&7.82&5.02&5.35&\textbf{4.50}\\
\bottomrule
\end{tabular}
\vspace{-0.5em}
\caption{\textbf{Quantitative evaluation on SceneFlow test set.} Our method achieves state-of-the-art performance on both metrics. \textbf{Bold}: Best.}
\vspace{-1.2em}
\label{tab:results_sceneflow}
\end{table*}

\subsection{Disparity Refinement}
\label{sec:refinement}
Thus far, the neural MRF inference on the coarse level features already exhibits competitive performance (\cref{tab:ablation_comp}). 
It can be further improved by performing $N_f$ neural message passing on the fine level features for detailed structure refinement. 
For efficiency, we use only one candidate label for every pixel on the fine level. 
The candidate label is obtained by reducing the coarse disparity estimation using a strided-4 median pooling with kernel size $4\times 4$. 
Because of a single candidate label, the refinement differs from coarse inference (\cref{sec:inference}) in two aspects: 1) the graph is free of \textit{self} edges, 2) posterior probability branch is no longer needed.

\subsection{Loss Functions}
\label{sec:loss}
To generate ground truth for proposal extraction at $\nicefrac{1}{8}$ resolution, we propose a superpixel-guided disparity map downsample operator, which reduces each non-overlapping $8\times 8$ patch to multiple modals (details in supplementary).

\paragraph{Proposal loss.} 
We expect candidate labels will identify all ground truth disparity modals $\{z_k^*\}$. 
To measure this, our loss finds an optimal bipartite matching between candidate labels and ground truth modals, and then optimizes the one-to-one matching. 
The candidate label set of pixel $o$, denoted by $L_o=\{\hat{z}_k\}$, is associated with proposal loss as:
\begin{equation}
    L^{\text{prop}}=\sum_k \text{Smooth}_{L1}\left(z_k^*-\hat{z}_{\hat{\sigma}(k)}\right)
    \label{eq:l_prop}
\end{equation}
where $\hat{\sigma}(k)$ is the index of candidate labels that has been matched with $z_k^*$. 
This loss is inspired by DETR~\cite{carion2020end}, a pioneering work in Transformer-based object detection.

\paragraph{Disparity loss.} Given the ground truth disparity $z^*$, the disparity loss is formulated as:
\begin{equation}
    L^{\text{disp}}=\sum_{z'}p(z')\vert z'-z^*\vert,
    \label{eq:loss_est}
\end{equation}
where $\{z'\}$ and $\{p(z')\}$ are the disparity hypotheses and posterior probabilities inferred by neural MRF.

%% file: 4_experiment.tex
\section{Experiments}

We evaluate the proposed stereo models on two widely used datasets, SceneFlow~\cite{mayer2016large} and KITTI~\cite{geiger2012we,menze2015object}. 
SceneFlow is a synthetic dataset which provides 35,454 training and 4,370 testing pairs of size $960\times 540$ with accurate ground truth disparity maps. 
KITTI 2012 and 2015 are real-world datasets with sparse LIDAR ground truth disparities for the training set. 
KITTI 2012 includes 194 training and 195 testing pairs, while KITTI 2015 has 200 training and 200 testing pairs. 
Additionally, we use the training pairs of KITTI, Middlebury 2014~\cite{scharstein2014high} and ETH3D~\cite{schops2017multi} to evaluate zero-shot generalization ability.

\subsection{Implementation Details}

We implement our model using the PyTorch framework on NVIDIA RTX 3090 GPUs. 
The AdamW optimizer is used in conjunction with a one-cycle learning rate scheduler for all training. 
On the SceneFlow dataset, we train models on $384\times 768$ random crops for 300k steps with a batch size of 8 and set the maximum learning rate to 0.0005. 
Following the standard protocol, we exclude all pixels with ground truth disparity greater than 192 from the evaluation. 
For our submissions to the KITTI benchmark, the model pre-trained on SceneFlow is fine-tuned on the mixed KITTI 2012 and 2015 training sets for another 39k steps with a batch size of 4.  
We randomly crop images to $304\times 1152$ and use a maximum learning rate of 0.0002 for fine-tuning. More implementation details are provided in the supplementary material.

\begin{table*}[!t]
\small
\centering
\begin{tabular}{c c c c c c c c c c}
\toprule
& \multicolumn{5}{c}{KITTI 2012} & \multicolumn{3}{c}{KITTI 2015} &\\
\cmidrule(lr){2-6}
\cmidrule(lr){7-9}
\textbf{Methods} & \multicolumn{2}{c}{\textbf{bad 2.0}} & \multicolumn{2}{c}{\textbf{bad 3.0}} & \textbf{EPE}  
& \textbf{BG} & \textbf{FG} & \textbf{ALL} & \textbf{Runtime}\\
\cmidrule(lr){2-3}
\cmidrule(lr){4-5}
& noc [\%] &  all [\%] &  noc [\%] &  all [\%] & noc [px] & \multicolumn{3}{c}{All Areas [\%]} & [s] \\
\cmidrule(lr){1-6}
\cmidrule(lr){7-10}
GCNet~\cite{kendall2017end} & 2.71 & 3.46  & 1.77 & 2.30 & 0.6 &  2.21 & 6.16 &  2.87 & 0.9\\
PSMNet~\cite{chang2018pyramid} & 2.44 & 3.01 & 1.49 & 1.89  & 0.5 & 1.86 & 4.62 &  2.32 & 0.41\\  
GwcNet~\cite{guo2019group} & 2.16 & 2.71 & 1.32 & 1.70 & 0.5 & 1.74 & 3.93 & 2.11 & 0.32\\
GANet-deep~\cite{Zhang2019GANet}  & 1.89 & 2.50 & 1.19 & 1.60 &\textbf{0.4}      &1.48 &3.46 & 1.81 & 1.8 \\
CSPN~\cite{cheng2019learning} & 1.79 & 2.27 & 1.19 & 1.53 & - & 1.51 & \underline{2.88} & 1.74 & 1.0\\
RAFT-Stereo~\cite{lipson2021raft} & 1.92 & 2.42 & 1.30 & 1.66 & \textbf{0.4} & 1.58 & 3.05 & 1.82 & 0.38\\
LEAStereo~\cite{cheng2020hierarchical} & 1.90 & 2.39 & 1.13 & 1.45 & 0.5 & 1.40 & 2.91 & 1.65 & 0.3\\
ACVNet~\cite{xu2022attention} & 1.83 & 2.35 & 1.13 & 1.47 & \textbf{0.4} & \underline{1.37} & 3.07 & 1.65 & 0.2\\
IGEV-Stereo~\cite{xu2023iterative} & 1.71 & \underline{2.17} & 1.12 & 1.44 & \textbf{0.4} & 1.38 & \textbf{2.67} & \textbf{1.59} & 0.18\\
PCWNet~\cite{shen2022pcw} & \underline{1.69} & 2.18 & \underline{1.04} & \underline{1.37} & \textbf{0.4} & \underline{1.37} & 3.16 & 1.67 & 0.44\\
\cmidrule(lr){1-6}
\cmidrule(lr){7-10}
PBCP~\cite{seki2016patch} & 3.63 & 5.01 & 2.36 & 3.45 & 0.7 & 2.58 & 8.74 & 3.61 & 68\\
LBPS~\cite{knobelreiter2020belief} & - & - & - & - & - & 2.85 & 6.35 & 3.44 & 0.39\\ 
Ours & \textbf{1.59} & \textbf{2.07} & \textbf{1.01} & \textbf{1.35} & \textbf{0.4} & \textbf{1.28} & 3.13 & \textbf{1.59} & 0.09\\
\bottomrule
\end{tabular}
\vspace{-0.5em}
\caption{\textbf{Quantitative evaluation on KITTI 2012 and 2015.} For KITTI 2012, we report the outlier ratio with error greater than $x$ pixels (bad $x$) in both non-occluded (noc) and all regions (all), as well as the overall EPE in non-occluded pixels. For KITTI 2015, we report D1 metric in background regions (BG), foreground regions (FG), and all. \textbf{Bold}: Best, \underline{Underline}: Second best.}
\label{tab:results_kitti}
\vspace{-1.2em}
\end{table*}

\subsection{Benchmark Evaluation}
We compare our stereo models with the published state-of-the-art methods on SceneFlow ``finalpass'', KITTI 2012 and KITTI 2015 datasets. 
The evaluation metrics on the SceneFlow test set are average end point errors (EPE) and bad pixel ratio with 1 pixel threshold (Bad 1.0). 
\Cref{tab:results_sceneflow} illustrates that our model achieves state-of-the-art performance on both metrics. 
Our method outperforms LEAStereo~\cite{cheng2020hierarchical} and ACVNet~\cite{xu2022attention} by 42.5\% and 10.4\% on the outlier ratio under 1 pixel threshold (Bad 1.0), respectively. 
These two are current state-of-the-art local methods~\cite{scharstein2002taxonomy} that regularize cost volume using 3D convolution networks. 

We also evaluate our method on the test set of KITTI 2012 and 2015, and the results are submitted to the \href{https://www.cvlibs.net/datasets/kitti/eval_stereo_flow.php?benchmark=stereo&table=all&error=2&eval=all}{online leaderboard}. 
At the time of writing, our method ranks \nth{1} on both datasets among all published methods while running faster than 100 ms. 
As shown in \cref{tab:results_kitti}, we achieve the best performance for almost all metrics on KITTI 2012 and 2015. 
Compared to prior leading global stereo methods, \eg, PBCP~\cite{seki2016patch} and LBPS~\cite{knobelreiter2020belief}, our method significantly outperforms them by more than 50\%. 
We present some qualitative results in \cref{fig:sceneflow_kt}. Our method performs well in large textureless and detailed regions.
In addition, we compare the point clouds generated by our approach and the top-performing LEAStereo~\cite{cheng2020hierarchical}. 
As depicted in \cref{fig:kitti-pt}, our approach can greatly alleviate \textit{bleeding artifacts}~\cite{tosi2021smd} and produce sharp disparity estimates near discontinuities. 

\subsection{Zero-shot Generation}
\begin{figure}[!t]
  \centering
  \tabcolsep=0.02cm
  \renewcommand*{\arraystretch}{0.25}
  \begin{tabular}{c c}
    \raisebox{2.1em}{\rotatebox{90}{\footnotesize Middlebury}} & \includegraphics[width=0.96\linewidth]{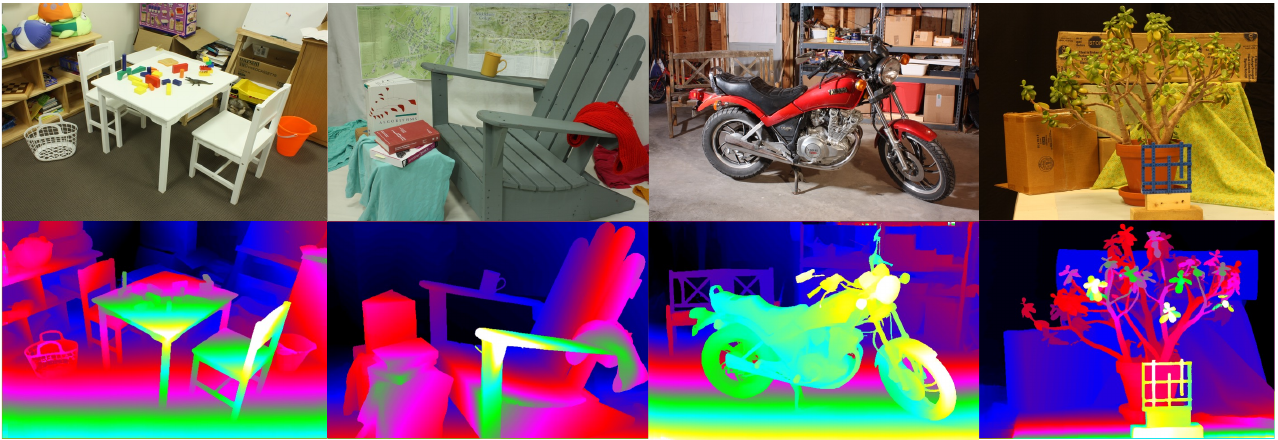} \\
    \raisebox{2.2em}{\rotatebox{90}{\footnotesize ETH3D}} & \includegraphics[width=0.96\linewidth]{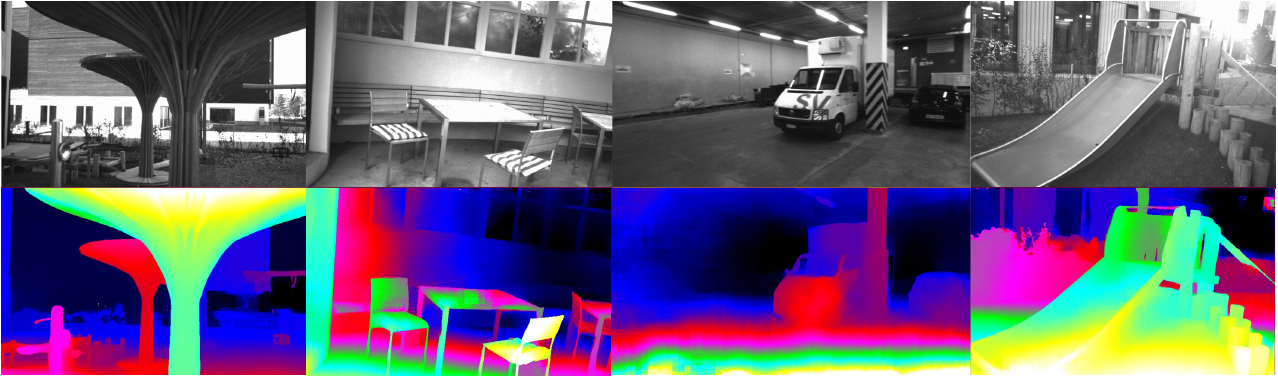}
  \end{tabular}
  \vspace{-0.7em}
  \caption{Zero-shot generalization on ETH3D and Middlebury.}
  \label{fig:qua-zero_shot}
  \vspace{-2.4em}
\end{figure}

Generalizing from synthetic training data to unseen real-world datasets is crucial because collecting large-scale real-world datasets for training is challenging and expensive. 
In this evaluation, we use the model trained only on synthetic SceneFlow dataset, \ie, that reports the accuracy of \cref{tab:results_sceneflow}, and directly test it on the KITTI, Middlebury and ETH3D training sets. 
\Cref{tab:results_zero_shot} compares our approach with current robust methods\footnote{IGEV-Stereo~\cite{xu2023iterative} uses a non-standard evaluation protocol. This paper reevaluates it using the standard protocol to compare with other methods.}. 
It turns out that our approach even surpasses DSMNet~\cite{zhang2019domaininvariant} and CREStereo++~\cite{jing2023uncertainty}, both of which are specifically designed for cross-domain generalization. 
\cref{fig:qua-zero_shot} shows some qualitive results for zero-shot generation on ETH3D~\cite{schops2017multi} and Middlebury~\cite{scharstein2014high}.

\begin{table}[!t]
\small
\centering
\tabcolsep=0.08cm
\begin{tabular}{c c c c c}
\toprule
Methods & KITTI-12 & KITTI-15 & Middlebury & ETH3D \\
\midrule
DSMNet~\cite{zhang2019domaininvariant} & 6.2 & 6.5 & \underline{8.1} & 6.2\\
RAFT-Stereo~\cite{lipson2021raft} & \underline{4.7} & 5.5 & 9.4 & \textbf{3.3}\\
CREStereo++~\cite{jing2023uncertainty} & \underline{4.7} & \underline{5.2} & - & 4.4\\
IGEV-Stereo~\cite{xu2023iterative} &5.2 & 5.7 & 8.8 & 4.0\\
\midrule
Ours & \textbf{4.2} & \textbf{5.1} & \textbf{7.5} & \underline{3.8}\\
\bottomrule
\end{tabular}
\vspace{-0.6em}
\caption{\textbf{Zero-shot generalization evaluation.} All methods are only trained on SceneFlow and evaluated based on the outlier ratio with error greater than the specific threshold. We use the standard thresholds: D1 for KITTI, 2px for Middlebury, 1px for ETH3D.}
\label{tab:results_zero_shot}
\vspace{-1.2em}
\end{table}

\subsection{Ablation Studies}
\paragraph{Effectiveness of disparity proposal network.} The upper limit of our architecture is determined by the quality of candidate labels. 
In this study, we evaluate the recall and accuracy of candidate labels. Two metrics are used: $x$-pixels recall, the percent of pixels with ground truth identified by candidate labels under the threshold $x$ pixels; and the average end point errors  (EPE) between the best label and ground truth, as visualized in ~\cref{fig:scene-robust}. 
\cref{tab:res_proposal} indicates that the DPN attains a recall of 99.8\% on popular datasets with 8-pixel threshold, which is equivalent to 1 pixel at the coarse level features. 
Moreover, the EPE metric of candidate labels is substantially lower than that of final estimation, \eg, 0.22 \vs 0.45 in SceneFlow. It implies that the DPN is not the performance bottleneck of our proposed architecture.

\begin{table}[!t]
\small
\centering
\begin{tabular}{c c c c c c}
\toprule
& \multicolumn{3}{c}{candidate labels} & \multicolumn{2}{c}{label seeds}\\
\cmidrule(lr){2-4}
\cmidrule(lr){5-6}
Datasets & 3px & 8px & EPE & 8px & 16px \\
& [\%]$\uparrow$ & [\%]$\uparrow$ & [px]$\downarrow$ & [\%]$\uparrow$ & [\%]$\uparrow$\\
\cmidrule(lr){1-4}
\cmidrule(lr){5-6}
SceneFlow-test & 99.29 & 99.77 & 0.22 & 91.94 & 94.85\\
KITTI-12 \textit{val} & 98.88 & 99.72 & 0.31 & 93.87 & 95.89\\
KITTI-15 \textit{val} & 99.66 & 99.95 & 0.21 & 93.91 & 95.92\\
\bottomrule
\end{tabular}
\vspace{-0.5em}
\caption{\textbf{Quantitative evaluation of disparity proposal.} 
The validation set of KITTI 2012 and 2015 both contain 40 image pairs. We report EPE and the $x$-pixel recall rate. The extracted candidate labels have significantly better quality than initialized label seeds.
}
\label{tab:res_proposal}
\vspace{-1.7em}
\end{table}

\begin{figure*}[!t]
  \centering
  \tabcolsep=0.02cm
  \begin{tabular}{c c}
  \raisebox{1.4em}{\rotatebox{90}{\footnotesize Ours \hspace{2.2em} LEAStereo \hspace{1.2em} Left Image}} & \includegraphics[width=0.96\linewidth]{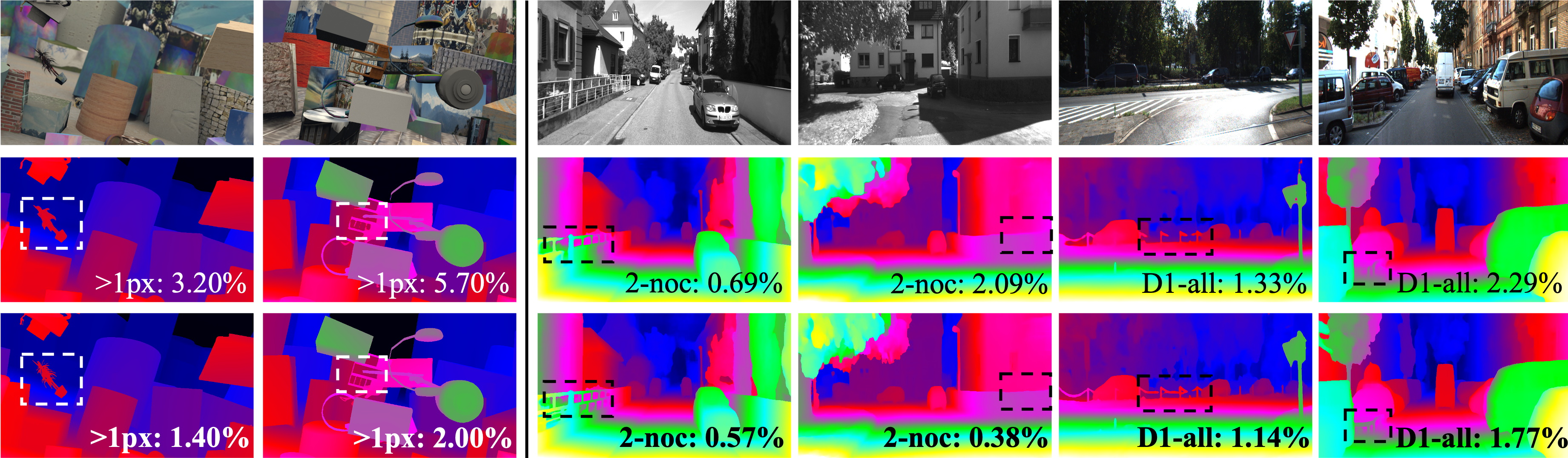}\\
  \end{tabular}
  \vspace{-0.8em}
  \caption{\textbf{Qualitative results on SceneFlow~\cite{mayer2016large} and KITTI~\cite{geiger2012we,menze2015object} benchmarks.} The leftmost two columns show results on SceneFlow, while the middle two and the rightmost two columns show results on KITTI 2012 and KITTI 2015, respectively. Our method exhibits outstanding performance in large textureless and detailed regions, compared with the top-performing LEAStereo~\cite{cheng2020hierarchical}.}
  \label{fig:sceneflow_kt}
  \vspace{-1.2em}
\end{figure*}

\begin{figure}[!t]
  \centering
  \vspace{0.4em}
  \includegraphics[width=0.96\linewidth]{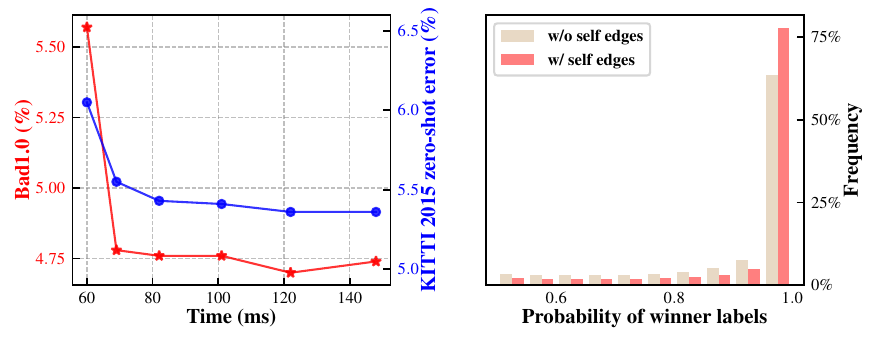}\\
  \vspace{-0.4em}
  \makebox[0.5\linewidth]{\hspace{1em}\footnotesize (a)}
  \makebox[0.02\linewidth]{}
  \makebox[0.45\linewidth]{\hspace{-1em}\footnotesize (b)}\\
  \vspace{-0.5em}
  \caption{(a) Tradeoff between time, accuracy, and generalization. 
  From left to right, the number of candidate labels are 1,2,3,4,5,6, respectively. (b) Histogram of probabilities of winner labels. Message passing along \textit{self} edges makes the winner more prominent. The statistics are based on 4.5M uniformly sampled pixels.
  }
  \label{fig:k_boxplot}
  \vspace{-1.0em}
\end{figure}

\paragraph{Number of candidate labels.} The results in \cref{tab:results_sceneflow,tab:results_kitti,tab:results_zero_shot,tab:res_proposal} are all achieved with the number of candidate labels $k=4$. 
We investigate the impact of the number of candidate labels on model accuracy, generalization ability, and inference time vary with. 
As shown in \cref{fig:k_boxplot}\textcolor{red}{a}, the proposed architecture is not sensitive to the number of candidate labels. 
Our method even achieves competitive performance with only one candidate label. 
Furthermore, our neural MRF is robust to occasional DPN failures (\cref{fig:scene-robust}). 
More candidate labels do not always result in better performance, since it increases the risk of choosing an incorrect hypothesis. As \cref{fig:k_boxplot}\textcolor{red}{a} suggests, $k=2$ may be a good choice for real time stereo matching with marginal sacrifice of performance.

\begin{table}[!t]
\vspace{1.8em}
\small
\centering
\begin{tabular}{l c c c}
\toprule
  & EPE[px] & Bad1.0[\%] & Time[s] \\
\midrule
baseline & 0.58 & 5.96& 0.064\\
w/ adaptive bias & 0.57 & 5.80 & 0.067\\
w/ position aggregation & 0.57 & 5.53 & 0.072\\
w/ \textit{shared} self edge & 0.56 & 5.48 & 0.072\\
w/ self edge & 0.53 & 5.24 & 0.081\\
\rowcolor{lightgray}
w/ refinement & 0.45 & 4.50 & 0.101\\
w/ refinement $\times 2$ & 0.43 & 4.13 & 0.121\\
\bottomrule
\end{tabular}
\vspace{-0.5em}
\caption{Ablation studies to investigate the effect of individual components on SceneFlow test. The baseline uses a fixed 2D relative positional bias~\cite{liu2021Swin}. The \nth{2} last row is our full model.}
\label{tab:ablation_comp}
\vspace{-1.8em}
\end{table}

\paragraph{Self edges.} Message passing along self edges is usually overlooked in hand-crafted stereo MRF and convolutional cost aggregation. 
This paper uncovered its significance in neural MRF inference. 
As shown in \cref{fig:k_boxplot}\textcolor{red}{b}, direct information exchange along self edges makes the winner label more prominent, such that the loss in \cref{eq:loss_est} is more focused on the disparity estimation of the winner label. 
This brings considerable improvements, \eg, lowering the EPE metric by 7\% (0.53 \vs 0.57), detailed in \cref{tab:ablation_comp}. 
We employ separate potential functions for self and neighbor edges. We validate this design by comparing it with the variant that uses the same function, denoted as \textit{shared} self edges in Tab.~\ref{tab:ablation_comp}. Our design consistently outperforms the alternative.

\paragraph{Attentional aggregation components.} We ablate adaptive bias and position aggregation in Tab.~\ref{tab:ablation_comp}. 
They both considerably reduce outliers. Adaptive positional bias performs better for edge pixels. We conjecture it brings benefits to the sharp depth boundaries shown in \cref{fig:kitti-pt}. 
By injecting positional encoding to \textit{value}, position aggregation results in superior estimation in large textureless regions (see \cref{fig:sceneflow_kt}).

\paragraph{Refinement.} Among all components in Tab.~\ref{tab:ablation_comp}, refinement contributes the most, as the fine level ($\nicefrac{1}{4}$) features would offer valuable information for detailed structures and pixels near boundaries. If we cascade two refinement modules together, better performance can be achieved with the compromise of generalization ability. This enables users to balance performance and generalization based on their needs.

\section{Conclusion}
We proposed a novel Neural MRF (NMRF) formulation for stereo matching and demonstrated its strong performance and generalization ability. 
It is distinct from the learning architectures currently in use, \eg, convolutional cost aggregation and iterative recurrent refinement.
We hope our new perspective will provide a new paradigm for stereo matching and can be extended to similar tasks, \eg, optical flow.

\paragraph{Acknowledgement.} This work is supported by Shenzhen Portion of Shenzhen-Hong Kong Science and Technology Innovation Cooperation Zone under HZQB-KCZYB-20200089, the RGC grant (14207320) from Hong Kong SAR government, CUHK T-Stone Robotics Institute, and the Hong Kong Center for Logistics Robotics.
\newpage

%% file: X_suppl.tex
\clearpage
\setcounter{page}{1}
\maketitlesupplementary

\setcounter{section}{0}
\section{Implementation Details of Loss Functions}
\paragraph{Superpixel-guided disparity downsample.}
To provide supervision signal for disparity proposal extraction at $\nicefrac{1}{8}$ resolution, we introduce a superpixel-guided disparity map downsample function, which reduces each $8\times 8$ disparity window to multiple modals. 
We divide the ground truth disparity map into non-overlapping $8\times 8$ windows and perform an independent downsample for each window.

\begin{figure}[!t]
\vspace{0.3em}
  \centering
  \includegraphics[width=0.98\linewidth]{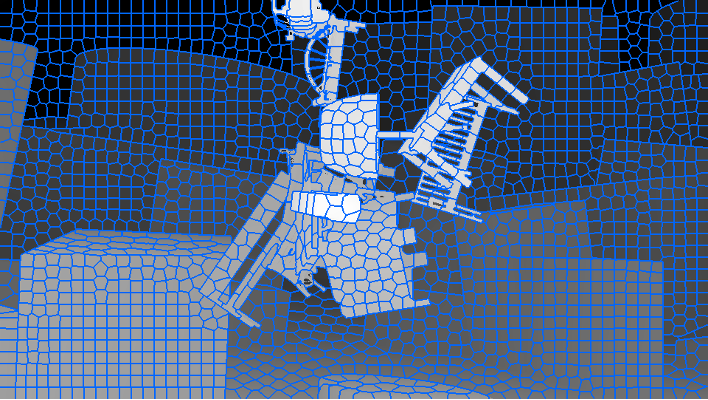}
  \caption{Superpixels overlaid on ground truth disparity map.}
  \label{fig:superpixel}
  \vspace{-1.2em}
\end{figure}

First, we over-segment the left image $I^L$ into superpixels using the LSC method implemented in OpenCV\footnote{\url{https://opencv.org}}.
As shown in \cref{fig:superpixel}, the superpixel effectively groups adjacent pixels while preserving local image structures, making it appropriate for reducing disparity values. 
Subsequently, each $8\times 8$ window is decomposed into multiple segments utilizing the superpixel label map. 
We sort the segments based on their pixel count and compute the \textit{median} disparity of each segment as the representative. 
To mitigate over-segmentation in the window, we employ a non-maximum suppression (NMS) on the representative disparity list. 
The suppression criterion is based on the difference between representative disparity values. 
If the absolute difference is less than 0.5 pixels, we merge the suppressed segment into the segment that suppresses it. 
After the merge step, we sort the segments again based on the pixel count and choose the median disparity of the top 4 segments as the downsample function output. If there exist fewer than 4 segments, we pad the output with null values.

\paragraph{Proposal loss.}
Once we have obtained the downsampled ground truth disparity modals, we use it to train our disparity proposal extraction network, as detailed in \cref{sec:loss} of the paper. 
When computing the proposal loss in \cref{eq:l_prop}, we need to find the optimal bipartite matching between proposals and ground truth modals. 
For instance, consider a pixel on the coarse level with four ground truth disparity modals, namely $\{1.1, 1.8, \phi, \phi\}$, and four extracted proposals, namely $\{1.4,10.2,10.8,11.2\}$. 
Ignoring null value $\phi$, the optimal bipartite matching pairs consist of $(1.1, 1.4)$ and $(1.8, 10.2)$. 
However, we need to be careful with the close ground truth modals. 
In this case, the proposal $1.4$ already captures the two close ground truth modals
$1.1$ and $1.8$. 
Thus, the matching pair $(1.8, 10.2)$ is unnecessary and may induce negative impact on the training. 

To address this, we perform an online non-maximum suppression (NMS) on ground truth modals when computing the proposal loss. 
First, we sort ground truth modals based on their proximity to the extracted proposal set. 
Proximity is measured by the minimum distance between the ground truth modal and all proposals. 
Then, we suppress the close ground truth modals using the threshold of $8$ pixels. 
In continuation with the above example, our online NMS reduces the ground truth modals to $\{1.1,\phi,\phi,\phi\}$, and only one matching pair $(1.1, 1.4)$ is leveraged for proposal loss. 

\paragraph{Initialization loss.} Besides the proposal loss, an initialization loss is also employed to supervise label seeds to identify ground truth modals. 
As described in \cref{sec:dpn}, label seeds are derived from a 3D cost volume $\mathbf{C}$, with
\begin{equation}
  \mathbf{C}(i,j,z)=\langle \tilde{F}^L(i,j),\tilde{F}^R(i-z,j)\rangle.
  \label{eq:3d-cost}
\end{equation}
We expect the initialization loss to penalize the discrepancy between ground truth modals and the 3D cost volume $\mathbf{C}$. 
To this end, we transform the ground truth modals of each pixel $o$ into a probability distribution, $p^*(z)=\sum_k w_k\delta(z-z_k^*)$, where $\{z_k^*\}$ are the ground truth modals at pixel $o$ and $\delta$ is the Dirac delta function. 
The mass weights $\{w_k\}$ are empirically set to $\{0.5,0.3,0.1,0.1\}$ for the four sorted ground truth modals. 
This simple strategy performs well in all our experiments. 
Note that ground truth modals are given with subpixel precision, however, label seed extractioin happens with integer disparities. 
Therefore, we displace the probability mass as $z_k^*$ to nearby integer disparities as
\begin{multline}
    \tilde{p}^*(z)=\sum_k w_k(\lfloor z_k^* \rfloor + 1 - z_k^*)\delta(z-\lfloor z_k^* \rfloor) + \\
    w_k(z_k^*-\lfloor z_k^* \rfloor)\delta(z-\lfloor z_k^* \rfloor-1).
\end{multline}
We define the initialization loss to be the cross entropy between ground truth probability $\tilde{p}^*$ and softmax of 3D cost volume $\mathbf{C}$ along the $z$ dimension, \ie,
\begin{equation}
    L^{\text{init}}=-\sum_{z\in [0,z_{\text{max}}]} \tilde{p}^*(z) \cdot \log(\text{softmax}_z(\mathbf{C}(i,j,z)).
\end{equation}

\section{Additional Implementation Details}
\paragraph{Local feature CNN.} We use a similar backbone as RAFT-Stereo~\cite{lipson2021raft}, which consists of a strided-2 stem and three residual blocks with strides 1, 2, 1, respectively. 
The network produces a feature map with 128 channels at $\nicefrac{1}{4}$ input image resolution, which is then downsampled through average pooling with a stride of 2 and a kernel size of 2. 
We further pass the obtained $\nicefrac{1}{8}$ resolution feature map and the original $\nicefrac{1}{4}$ resolution feature map to a shared convolution layer with 256 channels. 

\paragraph{Neural message passing.} The number of message passing blocks we use in label seeds propagation ($N_p$), MRF inference ($N_i$), and refinement ($N_f$) are 5, 10, 5 respectively. 
We use same settings for all experiments.
The channels of embedding vectors in all message passing blocks are always 128. 
The neighborhood window size is $4 \times 4$ for refinement (\cref{sec:refinement}), and $6 \times 6$ for neural MRF inference (\cref{sec:inference}). We also found that more message passing blocks and larger window size would bring slightly better accuracy with considerable computation overhead.

\paragraph{Observed label feature.} The observed feature of a candidate label must integrate matching cues from both left and right views. Given the coarse level features $\tilde{F}^L$ and $\tilde{F}^R$, we compute the observed feature $\mathbf{x}_v$ of a candidate label positioned at $\mathbf{p}_v:=(i,j,z)$ using a warping function \wrt $\tilde{F}^L(i,j)$ and $\tilde{F}^R(i-z,j)$ as
\begin{equation}
\begin{aligned}
  \mathbf{x}_v&=\text{MLP}\left(\mathbf{x}^{\text{concat}}_v || \mathbf{x}^{\text{corr}}_v\right)\\
  \mathbf{x}^{\text{concat}}_v&=\gamma_1\left(\tilde{F}^L(i,j)\right) || \gamma_1\left(\tilde{F}^R(i-z,j)\right)\\
  \mathbf{x}^{\text{corr}}_{v,g} &= \frac{N_g}{N_c}\langle \gamma_2\left(\tilde{F}^L_g(i,j)\right),\gamma_2\left(\tilde{F}^R_g(i-z,j)\right)\rangle,
\end{aligned}
\label{eq:obs_f}
\end{equation}
where $[\cdot\Vert\cdot]$ denotes concatenation along the channel dimension. $\tilde{F}_g^L,\tilde{F}_g^R$ are $g^{th}$ grouped features of $\title{F}^L$ and $\tilde{F}^R$, which are evenly divided into $N_g$ groups. 
$N_c$ is the channel of coarse level features, and $\langle\cdot,\cdot\rangle$ denotes the inner product. 
$\gamma_1$ and $\gamma_2$ are normalization functions to make the terms $\mathbf{x}_v^{\text{concat}}$ and $\mathbf{x}_v^{\text{corr}}$ share similar data distribution. 
Both $\gamma_1$ and $\gamma_2$ consist of two linear layers, with instance normalization and activation function following the first linear layer. 
Since disparity $z$ is a real number, we leverage bilinear interpolation when indexing feature map $\tilde{F}^R$. Our formulation is inspired by the success of GwcNet~\cite{guo2019group} and PCWNet~\cite{shen2022pcw}.

In the refinement stage, we use the same warping function as \cref{eq:obs_f}, but \wrt fine level features $\hat{F}^L$ and $\hat{F}^R$.

\begin{figure}[!t]
  \centering
  \begin{subfigure}[t]{0.48\linewidth}
  \includegraphics[width=\linewidth]{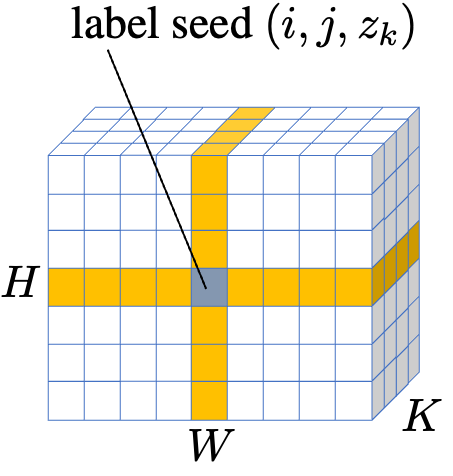}
  \caption{}
  \label{fig:cross-shaped}
  \end{subfigure}
  \begin{subfigure}[t]{0.48\linewidth}
    \includegraphics[width=\linewidth]{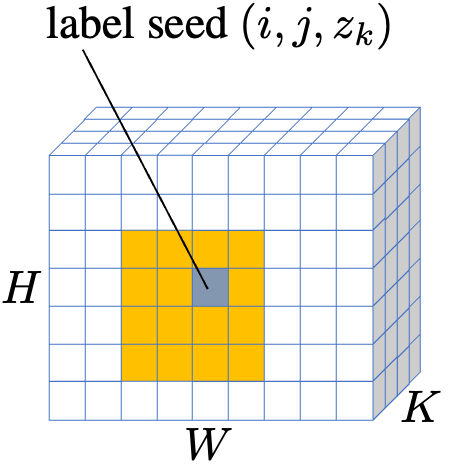}
    \caption{}
    \label{fig:local-win}
  \end{subfigure}
  \caption{(a) Cross-shaped attention window arrangement, (b) local attention window arrangement.}  
  \vspace{-1.2em}
\end{figure}

\paragraph{Cross-shaped window attention.} To efficiently capture long-range dependency for label seed message exchange, we employ the cross-shaped attention mechanism proposed in CSWin Transformer~\cite{dong2022cswin}. As illustrated in \cref{fig:cross-shaped}, the interested label seed, positioned at $\mathbf{p}_v:=(i,j,z_k)$, aggregates matching information from all other label seeds that share the same $i$ or $j$ coordinate. We follow the parallel multi-head grouping strategy and locally-enhanced positional encoding of CSWin Attention~\cite{dong2022cswin} when performing attentional aggregation. The initial matching feature $^{(0)}\mathbf{d}_v$ of a label seed $v$ is expected to encode cost features and underlying disparity value, formally defined as:
\begin{equation}
  ^{(0)}\mathbf{d}_v=\text{MLP}\Bigl(\gamma_3\left(L_z\left(\mathbf{C}(i,j,:)\right)\right)\Vert\ \text{PE}(z)\Bigr),
\end{equation}
where $\mathbf{C}$ denotes the 3D cost volume computed in label seeds extraction using \cref{eq:3d-cost}. The lookup operator $L_z$ retrieves cost features from volume slice $\mathbf{C}(i,j,:)$ around integer disparity $z$ for pixel $(i, j)$, akin to RAFT-Stereo~\cite{lipson2021raft}. We apply a two-layer MLP called $\gamma_3$ to normalize the retrieved cost features before concatenating it with the sinusoidal positional encoding (PE) of disparity $z$. 

We validate the design of cross-shaped window attention by comparing with the local window attention shown in \cref{fig:local-win}. The local window size is set to $8\times 8$ to match the computation complexity of cross-shaped window attention on SceneFlow dataset~\cite{mayer2016large}. The results are shown in \cref{tab:cswin}. Due to its ability to capture long-range dependency, cross-shaped window attention performs better than local window attention for label seed propagation. However, we do not adopt the cross-shaped window attention in neural MRF inference and refinement, since it does not adapt to our proposed content-adaptive positional bias and position aggregation for different input resolutions.

\begin{table}
  \small
  \centering
  \vspace{0.5em}
  \begin{tabular}{c c c c c c}
\toprule
& \multicolumn{2}{c}{candidate labels} & \multicolumn{2}{c}{disparity estimation}\\
\cmidrule(lr){2-3}
\cmidrule(lr){4-5}
 & 3px & 8px  & EPE & Bad 1.0 \\
& [\%]$\uparrow$ & [\%]$\uparrow$ & [px]$\downarrow$ & [\%]$\downarrow$ \\
\cmidrule(lr){1-3}
\cmidrule(lr){4-5}
cross-shaped & 99.29 & 99.77  & 0.45 & 4.50\\
local window & 99.18 & 99.72 & 0.47 & 4.56 \\
\bottomrule
\end{tabular}
\caption{Performance comparison between cross-shaped window attention and local window attention in label seed propagation.}
\label{tab:cswin}
\vspace{-1.2em}
\end{table}